\documentclass[conference]{IEEEtran}
\IEEEoverridecommandlockouts
\usepackage{cite}
\usepackage[caption=false]{subfig}
\usepackage{amsmath,amssymb,amsfonts}
\usepackage{algorithmic}
\usepackage{graphicx}
\usepackage{textcomp}
\usepackage{comment}
\usepackage{xcolor}
\usepackage{subfig}
\usepackage{float}
\usepackage{booktabs}
\usepackage{authblk}

\pdfoutput=1
\def\BibTeX{{\rm B\kern-.05em{\sc i\kern-.025em b}\kern-.08em
    T\kern-.1667em\lower.7ex\hbox{E}\kern-.125emX}}
\begin{document}

\title{Iterative Deep Learning Based Unbiased Stereology With Human-in-the-Loop\\
}

\author[1]{Saeed S. Alahmari}
\author[1]{Dmitry Goldgof}
\author[1]{Lawrence O. Hall}
\author[1]{Palak Dave}
\author[2]{Hady Ahmady Phoulady}
\author[ ]{ \\}
\author[3]{Peter R. Mouton}
\affil[1]{Department of Computer
Science and Engineering,
University of South Florida,
Tampa, FL, USA}
\affil[2]{Department of Computer Science,
 University of Southern Maine, 
 Portland, ME, USA}
 \affil[3]{SRC Bioscience\\
Tampa, FL USA}
 \affil[ ]{\textit{\{saeed3,goldgof,lohall,palakdave\}@mail.usf.edu, \{hady.ahmadyphoulady\}@maine.edu,}}
 \affil[ ]{\textit{and \{peter\}@disector.com}}

\renewcommand\Authands{ and }
\maketitle

\begin{abstract}
Lack of enough labeled data is a major problem in building machine learning based models when the manual annotation (labeling) is error-prone, expensive, tedious, and time-consuming.  In this paper, we introduce an iterative deep learning based method to improve segmentation and counting of cells based on unbiased stereology applied to regions of interest of extended depth of field (EDF) images. This method uses an existing machine learning algorithm called the adaptive segmentation algorithm (ASA) to generate masks (verified by a user) for EDF images to train deep learning models. Then an iterative deep learning approach is used to feed newly predicted and accepted deep learning masks/images (verified by a user) to the training set of the deep learning model. The error rate in unbiased stereology count of cells on an unseen test set reduced from about 3 \% to less than 1 \% after 5 iterations of the iterative deep learning based unbiased stereology process.   
\end{abstract}


\section{Introduction}
Unbiased stereology is a set of theoretical and practical methods for making accurate counts of stained cells by carefully avoiding all known sources of methodological bias \cite{mouton2011unbiased}\cite{mouton2017unbiased}. Examples of common stereology parameters include counts of total cell number and cell density; region and mean cell volumes; surface area and surface density; and total length and length density \cite{burke2009knowing}\cite{mouton2011unbiased}. However, current computer-assisted stereology systems available to bioscientists and medical scientists are based on a technology developed more than two decades ago. Therefore, a simple study requires tedious counting of hundreds of cells per sample by a well-trained technician \cite{west1991unbiased}. For example, a simple count of immunostained cells in a defined region of interest (ROI) requires about 2-3 {\bf hours} for a well-trained technician to achieve a reliable result. Though based on theoretically unbiased principles, this approach is prone to data errors and low reproducibility due to user subjectivity, variable expertise, and fatigue. The Adaptive Segmentation Algorithm (ASA) \cite{mouton2017unbiased} makes stereology counts of total numbers of brain cells (Neu-N immunostained neurons) by automatic segmentation and cell counting on Extended Depth of Field (EDF) images \cite{valdecasas2001extended}\cite{bradley2004one}. However, ASA requires manual adjustment of several parameters (i.g., minimum cell size, cell maximum size, and Gaussian Mixture Model (GMM) threshold) to achieve a good result. In Section ~\ref{sec:ASA}, we present  ASA details.
\newline \\
\indent A critical aspect of building successful machine learning models is the availability of labeled data. However, labeled data is hard to obtain because the process is time-consuming, labor-intensive, and tedious. Additionally, data labeling in a medical field is mostly restricted to experts in the field and generally cannot be prepared by a crowdsourcing approach for reasons such as the quality of annotation and subject privacy. To overcome labeling difficulties for stereology images (i.e., creating pixel-wise labels), we propose an iterative deep learning method to generate segmentation masks of cells on stained NeuN tissue images; then a human-in-the-loop approach was taken to verify each predicted mask and feed correct images-masks pairs to the training set.     
\newline \\
\indent Deep neural networks have generated considerable interest in the medical imaging field because they have shown performance advantages over conventional engineered image analysis algorithms. Although the idea of neural networks has been around for a long time, the recent deep neural networks revolution is partly due to the development of the convolutional neural network (CNN), optimization algorithms \cite{hinton2006reducing} \cite{nair2010rectified} \cite{srivastava2014dropout} \cite{ioffe2015batch}, and powerful, efficient computation resources. Deep learning refers to learning methods that often start from raw data get to a more abstract level \cite{lecun2015deep}. Convolutional Neural Networks have shown significant success in challenging tasks in image classification and recognition \cite{litjens2017survey} \cite{krizhevsky2012imagenet}. In this paper, we use a CNN based architecture for medical image segmentation known as Unet \cite{ronneberger2015u}. This architecture is a simple, fast, and end-to-end fully convolutional network that contains contraction and expansion paths to capture context and learn precise localization.
\newline \\
\indent In this paper, we propose a method that iteratively utilizes deep learning with human-in-the-loop and an existing unsupervised algorithm (ASA) which eliminates human data labeling entirely (creating masks) of NeuN stained images to quantify the number of cells in an ROI. This approach uses a state-of-art deep learning architecture in which user verified ASA results of EDF images are used to train a convolution neural network (CNN) model to segment and make automatic neuron counts on test images. Meanwhile, a set of deep learning predicted masks are verified by a human-in-the-loop and fed back to the train set. The main innovation is: i) elimination of human labeling effort (creating masks) using an existing unsupervised algorithm (ASA) to generate masks to train a CNN, ii) using a deep learning iterative process to reduce human effort in data labeling, where the user only verifies the correctness of segmentation, and iii) improving deep learning stereology cell counting by adding correctly labeled images (EDF images and their corresponding masks) to the training set for the next iteration.
\section{Unbaised Stereology}
\indent Unbiased stereology is the state-of-the-art for biological objects quantification in tissue sections\cite{saper1996any}. An essential component of this approach is unbiased sampling (i.e., systematic-random) that avoids all sources of biased assumption such as shape, size, and orientation \cite{saper1996any} \cite{burke2009knowing}. Unbiased stereology uses a virtual disector box to quantify the number of cells in a region-of-interest (ROI). Counted cells are based on their location within an ROI and disector box. For instance, cells touching the disector inclusion-line (i.e., disector upper and right line) or inside the disector box are counted. However, cells that touch the exclusion line (i.e., disector lower and left line) are not counted. An example of the disector box counting procedure is shown in Fig. ~\ref{fig:Annotation}, where the green line represents inclusion line, and the red line represents the exclusion line. Counted cells are marked manually with the blue marks.
\section{Data set} 
\label{sec:dataset}
The data set used in this work was sampled from the neocortex brain region of Tg4510 mice. As described by Mouton et al. in \cite{mouton2017unbiased}, animals and the process used in this study were approved by the University of South Florida (USF) Institutional Animal Care and Use Committee which follows NIH guidelines. The data set includes both genetically modified mice and control mice. Mice neurons change while expressing mutant tau. These neuron changes include neuron degeneration and neuroglia cells activation \cite{mouton2017unbiased}\cite{santacruz2005tau}\cite{spires2006region}. Mice samples were stained with NeuN single staining from which counting was performed manually using an optical fractionator \cite{west1991unbiased}. Disector stacks were captured and saved using the Stereologer system \cite{mouton2017unbiased}. Table ~\ref{LU_info} shows the number of sections from which multiple stacks were obtained and converted into EDF images. The total number of EDF images we have is 966 with their corresponding ASA masks.

\begin{table}[ht]
\centering
\caption{Datasets mouse ID, number of sections per mouse and total number of stacks per mouse}
\label{LU_info}
\begin{tabular}{@{}c|c|c@{}}
\toprule
Mouse ID  & Number of sections & Number of stacks \\ \midrule
02            & 8                  & 113              \\
03            & 6                  & 121              \\
14          & 8                  & 90              \\
17          & 7                  & 91               \\
29          & 8                  & 135              \\ 
21 	  & 7				   & 102			  \\
24 		  & 8				   & 103			  \\
67 		  & 8				   & 104			  \\
09 		      & 6				   & 107			  \\
\bottomrule
\end{tabular}
\end{table}
\section{Adaptive Segmentation Algorithm}
\label{sec:ASA}
\begin{figure}[htbp]
\centering
\subfloat[][Manual annotation]{\includegraphics[width=0.4\linewidth]{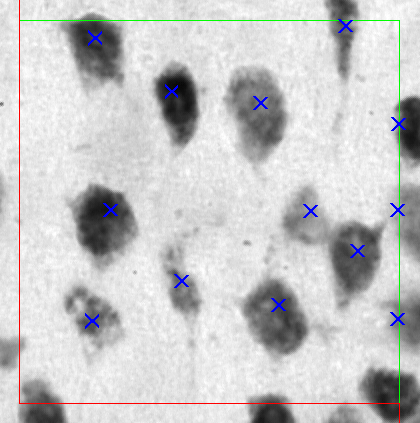}\label{fig:Annotation}}
   \qquad
\subfloat[][EDF image]   
{\includegraphics[width=0.4\linewidth]{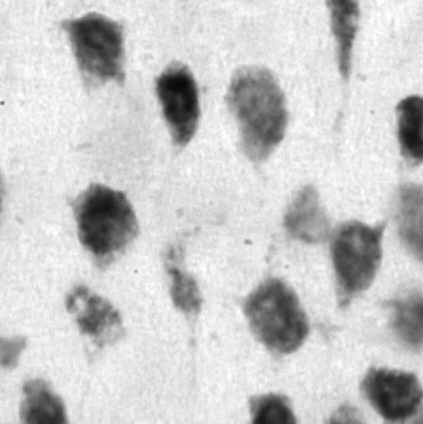}\label{fig:EDF_image}}

\subfloat[][ASA mask]   
{\includegraphics[width=0.4\linewidth]{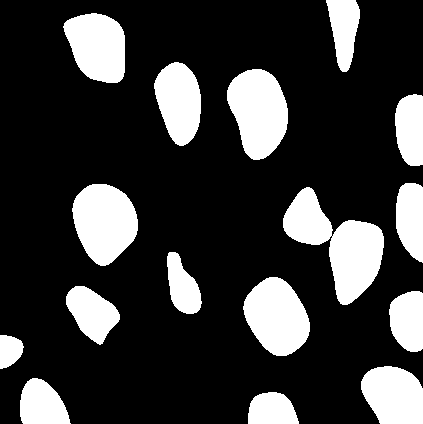}\label{fig:ASA_mask}}
\caption{An example from our data set, where a) is the manual  annotation (counted neurons have green dots), b) is the EDF image, and c) is the ASA mask for the EDF image shown in (b)}
     \label{fig:images_vis}
\end{figure}

As shown in \cite{mouton2017unbiased}, the adaptive segmentation algorithm (ASA) consists of multiple steps optimized to segment cells at high magnification (63 to 100x oil immersion) microscopy. The ASA includes a Gaussian Mixture (GMM), morphological operations, Voronoi diagrams, and watershed segmentation. It starts with EDF images to segment NeuN stained cells within a region of interest (ROI) using GMM; where GMM uses pixel intensity for the Expectation Maximization algorithm (EM) to estimate its components followed by thresholding and morphological operations to get separate cells. A processed EDF image using opening then closing by reconstruction is used in the watershed foreground and background markers extraction. These foreground and background markers are used in applying the watershed segmentation followed by segmentation approximation using  Voronoi diagrams. ASA uses a smoothing process to enhance cell boundaries using a Savitzky-Golay filter \cite{savitzky1964smoothing}. The reason to use ASA  is that manual annotation does not provide mask information, but instead, it provides a mark of what cell is being counted based on the unbiased stereology approach. An example of manual annotation is shown in Fig. ~\ref{fig:Annotation}.

\begin{figure}[htbp]
\centering
\subfloat[][Initial masks created using ASA followed by human verification]{\includegraphics[clip, trim=2cm 5cm 2cm 4cm,width=\linewidth]{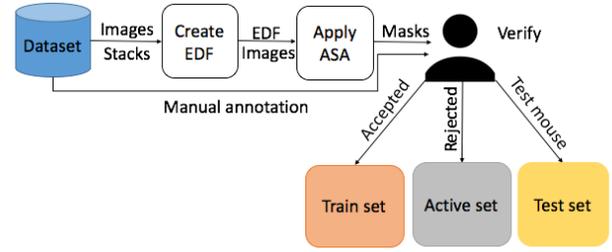}\label{fig:initalProcess}}
   \qquad
\subfloat[][Iterative human-in-the-loop verification of deep learning predicted masks]   
{\includegraphics[clip, trim=2cm 3cm 2cm 2cm,width=\linewidth]{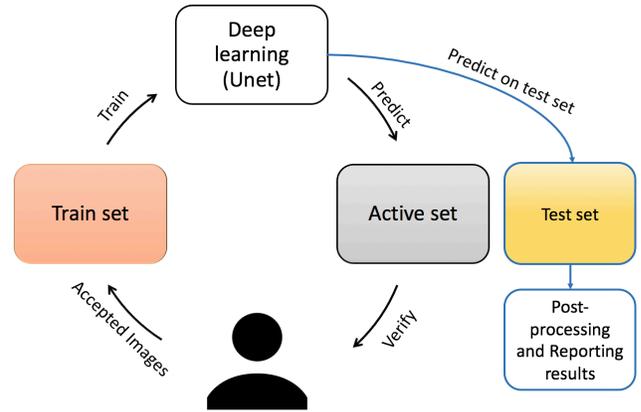}\label{fig:Human_in_loop}}

\caption{Proposed method in two steps: a) creating EDF images, and applying ASA, then human verification, and then b) iterative process using accepted ASA masks/images for training, and ASA masks/images as an active set. Human verification (i.e., accept or reject) on every predicted mask. Test set is a separate mouse (mouse id 17) }
     \label{fig:proposedDiagram}
\end{figure}
\section{Method}
\label{sec:method}
 In unbiased stereology, where labeling cells is tedious, time-consuming, and subject to errors, an iterative deep learning approach can leverage the data labeling process and generate correctly labeled examples that could help in building a more robust model. The EDF algorithm \cite{bradley2004one} was applied to each stack of images to produce a single in-focus image as shown in Fig. ~\ref{fig:EDF_image}. Using EDF images, initial labels (masks) for our data set have been created using the ASA algorithm as shown in Fig. ~\ref{fig:ASA_mask},  followed by a manual verification step to identify ASA accepted masks/images (i.e., ASA masks which match the manual annotation) from ASA rejected masks/images (i.e., ASA masks do not match the manual annotation) as illustrated in Fig. ~\ref{fig:initalProcess}. An example of manual annotation marks is shown in Fig. ~\ref{fig:Annotation}, where blue marks correspond to counted cells. For instance, the image shown in Fig. ~\ref{fig:ASA_mask} was accepted by a user because every cell mask (white blobs) inside or touching inclusion line (i.e., upper and right green line shown in Fig. ~\ref{fig:Annotation}) correspond to blue marks (counted cells) in the manual annotation shown in Fig. ~\ref{fig:Annotation}. This user verification step ignores cells that touch the exclusion line (disector left and lower line) for the purpose of training deep learning model.  
\newline \\
\indent In this section, we describe our iterative deep learning approach which can be described in five steps: 1) we train a deep learning model (Unet) on EDF images, and their corresponding ASA accepted masks that match manual annotation images; 2) A prediction was made on EDF images of ASA rejected masks that do not match manual annotation, and we refer to this set of images as the "active set";  3) Another set that does not overlap with either the training nor the active set is the "test set" which contains EDF images of different sections of a unique mouse (mouse id 17); 4) The results of testing a trained deep learning model on the active set were verified by the user by comparing the predicted mask and manual annotation similarity (as described in the previous paragraph). If a mask matches the manual annotation image (i.e., cells marked for counting on a manual annotation image were segmented correctly using deep learning), then that mask and its corresponding EDF image are augmented using a combination of elasticity and rotation augmentation and then the augmented images are added to the training set for deep learning. Meanwhile, the EDF gets removed from the active set. On the other hand, if a mask does not match with its corresponding manual annotation image, then its EDF image remains in the active set; 5) The iterative process was performed for five iterations. Fig. ~\ref{fig:Human_in_loop} demonstrates the proposed method. It is important to note that human-in-the-loop involvement means accepting or rejecting a mask based on its corresponding manual annotation as shown in Fig. ~\ref{fig:Annotation}. Therefore, no relabeling was made by the human-in-the-loop.
\newline \\
\indent For each iteration of our method, we trained a deep learning architecture (Unet) for 100 epochs using Keras and Tensorflow backend \cite{chollet2015keras}\cite{tensorflow2015-whitepaper}. The Adam optimizer was used where the learning rate was set to $1e^{-4}$, while exponential decay rates for the moment estimates hyperparameters $\beta 1$ (first moment) and $\beta 2$ (second moment) were set to 0.9 and 0.999 respectively. \cite{kinga2015method}.
\newline \\
\indent Based on the unbiased stereology method, cells counted in an ROI are those stained cells that are located inside the ROI or touching the inclusion line (i.e., top and right green line) but not touching the exclusion line (i.e., bottom and left red lines) as shown in Fig. ~\ref{fig:Annotation}. For training purposes, we have kept all cells by ignoring the unbiased stereology constraints. However, prior to reporting the results on the test set (mouse id 17), a postprocessing step was applied to remove small noise on the predicted mask, separate touching cells, and to impose unbiased stereology criteria of counting by removing cells that touch the exclusion disector line.

\section{Experiments and Results}
\label{sec:Exp_and_results}
Our data set has 966 stacks from 9 different mice. The EDF algorithm was used to create EDF images for each stack to get an in-focus image. The number of images in the initial train set (no augmentation) is 147 images, the number of images in the initial active set is 728 images, and the number of images in the test set is 91 images. Data augmentation used in this experiment was rotation 15\textdegree {} of elastic augmentation \cite{simard2003best}. For example, for an image $\textit{M}$, we apply an elastic algorithm with two different random seeds, which yields two elastic images $\textit{M}_1$, $\textit{M}_2$. Then for each of $\textit{M}$,  $\textit{M}_1$, and $\textit{M}_2$ we apply rotation augmentation of 15\textdegree. The total number of images generated by applying elastic then rotation augmentation of a single image is 72 images (including original image). This elastic and rotation augmentation is applied to the EDF images, and then we cropped images 20 pixels around the disector line as shown in Fig.  ~\ref{fig:EDF_image}. We have used error rate to report results on the test set as shown in Equation ~\ref{equation:ErrorRate}, where $y_{true}$ is the number of counted cells on ground truth (manual annotation), and $y_{pred}$ is the number of counted cells on a predicted deep learning mask. For iteration 1, training on ASA accepted only images, and testing on a separate mouse (i.e., test set) resulted in 3.16 \% error rate, and the user accepted 379 images from the active set. Increasing training images helped to reduce the error rate on the second iteration to 0.82 \%; furthermore, 81 images were accepted by the user and moved to the training set. The lowest error rate on the test set was 0.41 \% with a higher number of training images (iteration 4).    
\begin{equation}
Error\ rate = \frac{\left|y_{true} - y_{pred}   \right|}{y_{true}}* 100
\label{equation:ErrorRate}
\end{equation}  
\newline

\begin{table}[!htp]
\centering
\caption{Results of the proposed method that shows the number of accepted images from active set in every iterartion, and the error rate (\%) on a test set (Mouse id 17)}
\label{table:results}
\resizebox{0.48\textwidth}{!}{%
\begin{tabular}{@{}c|c|c@{}}
\toprule
Iteration number & Number of accepted images & Error rate on test set (\%) \\ \midrule
1                & 379                        & 3.16                     \\
2                & 81                        & 0.82                    \\
3                & 51                        & 1.92                     \\
4                & 18                         & \textbf{0.41}                     \\
5                & 15                         & 0.55                     \\ \bottomrule
\end{tabular}%
}
\end{table}


\begin{figure}[ht]
\centering
\subfloat[][ASA]{\includegraphics[width=0.4\linewidth]{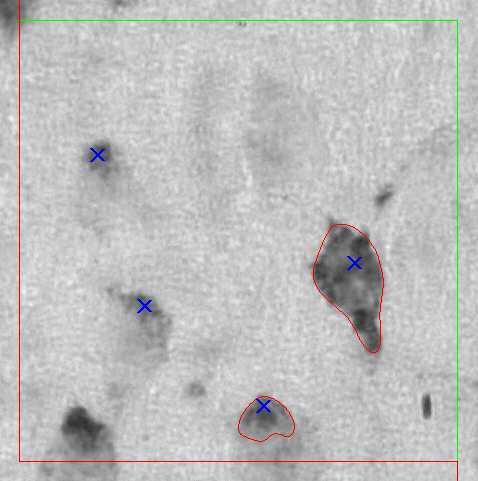}\label{fig:ASA}}
\qquad
\subfloat[][deep learning]   
{\includegraphics[width=0.4\linewidth]{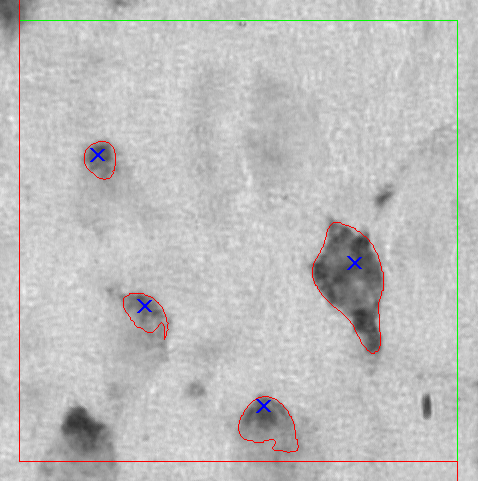}\label{fig:DeepLearning}}
\caption{Example from our data set, where a) the ASA masks contour overlaid on manual annotation image (counted neurons have blue marks), b) the iterative deep learning predicted masks contour (accepted  on the fifth iteration of our iterative deep learning based unbiased stereology) overlaid on manual annotation image}
\label{fig:images_vis2}
\end{figure}

\begin{figure}[htbp]
\centering
\includegraphics[width=1\linewidth]{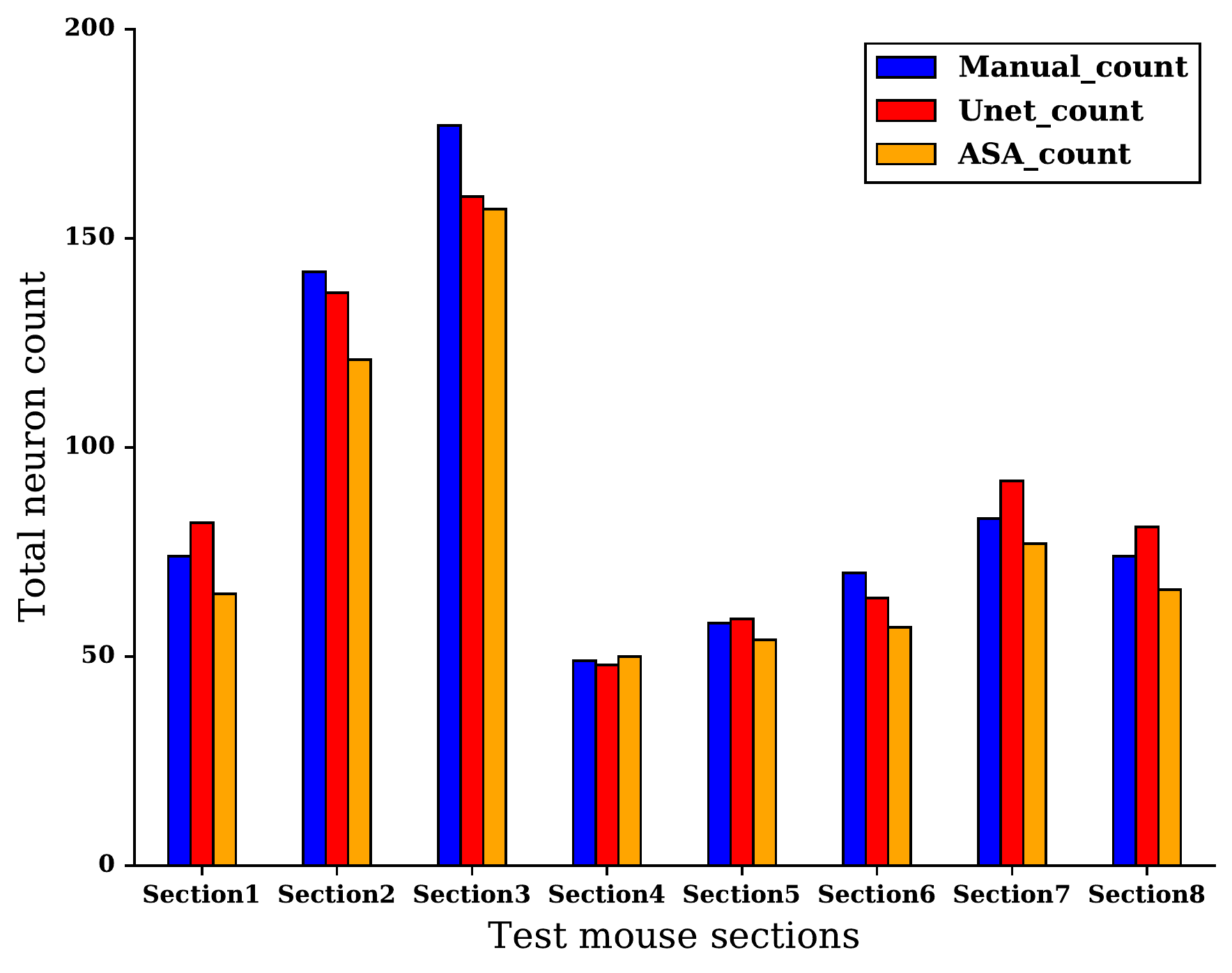}
\caption{Test mouse cells count using manual, ASA, and Unet (deep learning)}
\label{fig:barChart}
\end{figure}

\begin{table}[!htp]
\centering
\caption{Test mouse cells count using manual method, Unet (deep learning), and ASA.}
\label{table:LU-17}
\begin{tabular}{@{}l|c|c|c@{}}
\toprule
Test mouse & Manual cells count & Unet cells count & ASA cells count \\ \midrule
Section 1     & 74            & 82          & 65        \\
Section 2     & 142             & 137           & 121         \\
Section 3     & 177            & 160          & 157         \\
Section 4     & 49            & 48          & 50         \\
Section 5     & 58           & 59         & 54        \\
Section 6     & 70            & 64          & 57         \\
Section 7     & 83            & 92         & 77 
\\
Section 8     & 74            & 81         & 66
\\ \bottomrule
\end{tabular}
\end{table}

\begin{figure}[ht]
\centering
\subfloat[][ASA]{\includegraphics[width=0.4\linewidth]{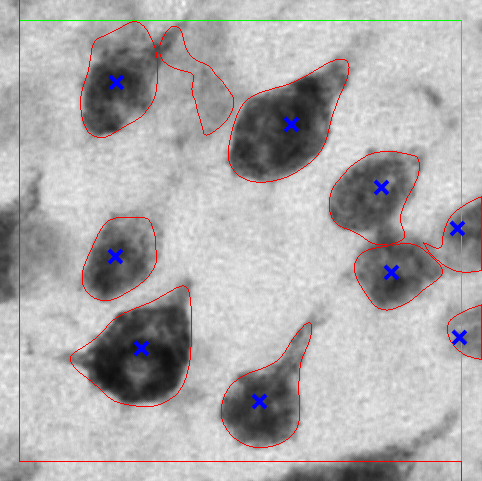}\label{fig:ASA_5a}}
\qquad
\subfloat[][Deep learning]   
{\includegraphics[width=0.4\linewidth]{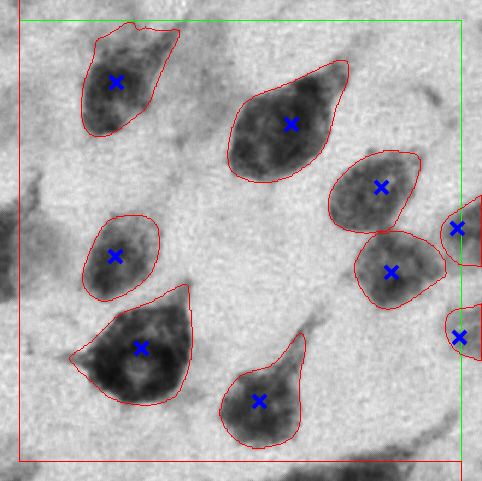}\label{fig:DeepLearning_5b}}
\\
\subfloat[][ASA]{\includegraphics[width=0.4\linewidth]{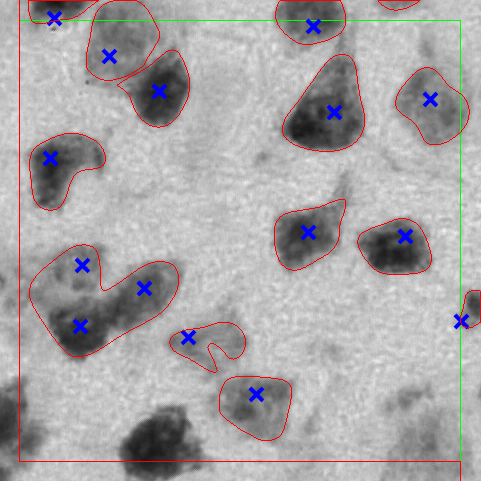}\label{fig:ASA_5c}}
\qquad
\subfloat[][Deep learning]   
{\includegraphics[width=0.4\linewidth]{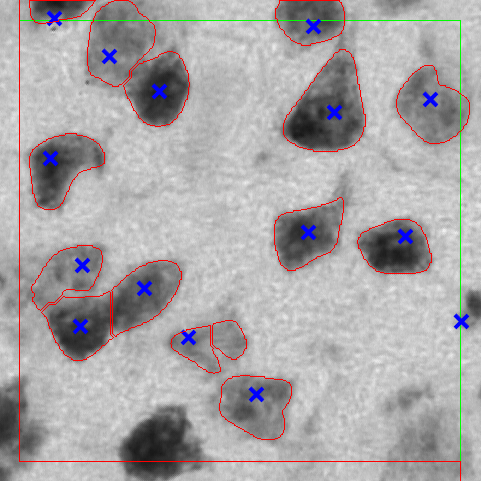}\label{fig:DeepLearning_5d}}
\\
\subfloat[][ASA]{\includegraphics[width=0.4\linewidth]{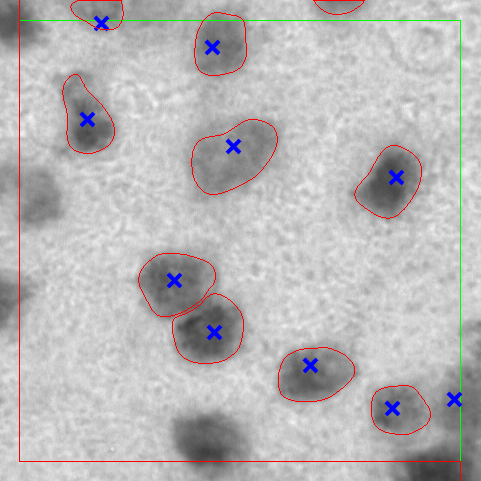}\label{fig:ASA_5e}}
\qquad
\subfloat[][Deep learning]   
{\includegraphics[width=0.4\linewidth]{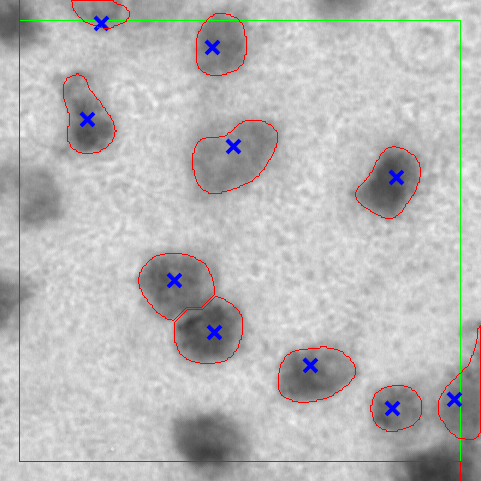}\label{fig:DeepLearning_5f}}

\caption{Examples from the test set, where a,c, and d are the ASA mask contours overlaid on manual annotation images (counted neurons have blue marks), b, d, and f are the iterative deep learning predicted masks (iteration 5) contours overlaid on manual annotation image}
\label{fig:images_vis3}
\end{figure}

In Table ~\ref{table:results}, complete results of the iterative deep learning based unbiased stereology approach of five iterations are provided. Fig. ~\ref{fig:images_vis2} shows an example from our data set to compare the ASA mask, and the iterative deep learning predicted mask. Fig. ~\ref{fig:DeepLearning} shows an improved segmentation of cells on EDF image which was accepted on the fifth iteration. One unanticipated finding was that on the third and fifth iterations, the error rate was slightly higher than the prior iteration. This could be caused by the rotation augmentation artifacts and user subjectivity when accepting new images/masks from the active set. In Fig. ~\ref{fig:barChart}, we show a visual comparison between the iterative deep learning iteration (iteration 4) results, ASA, and manual counting. Where manual cell count, ASA based cell count, and deep learning (Unet) based cell count are reported for the test set. The ASA error rate on the test set (mouse id 17) was 11 \%. Additionally, the cell count of the test mouse in different sections is shown in Table ~\ref{table:LU-17}. A comparison between ASA results and Unet (Deep learning) results on three examples from the test set after post-processing and applying unbiased stereology counting rules is presented in Fig. ~\ref{fig:images_vis3}.
\newline

\section{Discussion}
The evidence from this study suggests that the iterative deep learning based unbiased stereology method presented herein is much faster and more accurate than the state-of-the-art stereology since human involvement was mainly reduced. The state-of-the-art stereology takes about 2-3 hours per an ROI; however, the proposed method herein estimated time was approximately  20-30 minutes per an ROI (including preparing masks using ASA, human verification, using the trained model for the prediction on the test set, post-processing, and counting). Human involvement reduction was as follows: 
\begin{enumerate}
  \item Instead of creating initial labels manually (creating masks), an unsupervised algorithm (ASA) was utilized to create initial masks, then a user verification step to merely accept or reject an image/mask based on the match to the manual annotation.
  \item Instead of relabeling active set images/masks predicted by deep learning in each iteration of the iterative deep learning process, the human-in-the-loop was only verifying the correctness of predicted masks, that is accepting or rejecting based on the match to the manual annotation as described earlier.
\end{enumerate}
\vspace{3mm}

\indent Lack of a large number of properly annotated images remains an obstacle to many researchers, especially in the medical field; because manual image annotation is tedious and error-prone work. Additionally, lack of expert labor to annotate a large number of images to build a more robust model is an issue. Therefore, utilizing a pre-existing algorithm (i.e., ASA) to generate masks for EDF images overcomes significant challenges such as creating masks manually. However, a user intervention in creating masks was not eliminated but instead reduced; since ASA masks need a verification step to accept masks that match manual annotation. As a result, training of the deep learning model was done on EDF images, and their corresponding accepted ASA masks. Therefore, it can thus be suggested that using pre-existing methods such as ASA for initial mask generation followed by a verification step is the most suitable way to accelerate data labeling (i.e., segmentation)  that would have taken a large amount of time otherwise. Additionally, using a human-in-the-loop for iterative deep learning to verify predicted masks on an unlabeled set of EDF images (i.e., active set) is an optimal method to increase training images with correct labels (masks). \\

\indent The generalisability of this study is subject to certain limitations. For instance, lack of enough data to best train deep learning models. Another limitation is user subjectivity in verifying predicted masks by deep learning in the active set. Notwithstanding the relatively limited data and user subjectivity constraints, this work offers valuable insights into using an existing unsupervised algorithm (ASA) to generate masks (labels) instead of human labeling (creating masks), then improving the model performance by iterative deep learning based unbiased stereology with a human-in-the-loop.

\section{Conclusion}
This paper presents an iterative deep learning based unbiased stereology strategy that uses an existing unsupervised algorithm (ASA) to obtain masks as initial labels for training a deep convolutional neural network to segment and count cells on ROIs of NeuN single stained images. The proposed method herein was able to achieve good results (error rate less than 1 \%) compared to the ASA cell counting (error rate of 11\%), although ASA generated the initial labels (segmentation masks).  Moreover, the proposed algorithm eliminates human effort in data labeling, where human-in-the-loop work was merely to verify masks based on the corresponding manual annotation. Our approach has some drawbacks such as human-in-the-loop effort and subjectivity which could be an obstacle with massive sets of images for verification. Iterative deep learning based unbiased stereology techniques in conjunction with initial labels (masks) from an existing algorithm (ASA) showed encouraging results compared to ASA and the current stereology counting method which is laborious, slow, and time-consuming to obtain for a significant amount of data.    
\newline

\section{Acknowledgments} 
The authors would like to thank Dr. Marcia Gordon of Michigan State University (Grand Rapids, MI) for the generous donation of the stained tissue sections for these studies. This research was partially supported by the National Science Foundation under award number (\#1513126) and (\#1746511). 
 
\bibliographystyle{IEEEtran}
\bibliography{conference_041818}

\begin{thebibliography}{10}
\providecommand{\url}[1]{#1}
\csname url@samestyle\endcsname
\providecommand{\newblock}{\relax}
\providecommand{\bibinfo}[2]{#2}
\providecommand{\BIBentrySTDinterwordspacing}{\spaceskip=0pt\relax}
\providecommand{\BIBentryALTinterwordstretchfactor}{4}
\providecommand{\BIBentryALTinterwordspacing}{\spaceskip=\fontdimen2\font plus
\BIBentryALTinterwordstretchfactor\fontdimen3\font minus
  \fontdimen4\font\relax}
\providecommand{\BIBforeignlanguage}[2]{{%
\expandafter\ifx\csname l@#1\endcsname\relax
\typeout{** WARNING: IEEEtran.bst: No hyphenation pattern has been}%
\typeout{** loaded for the language `#1'. Using the pattern for}%
\typeout{** the default language instead.}%
\else
\language=\csname l@#1\endcsname
\fi
#2}}
\providecommand{\BIBdecl}{\relax}
\BIBdecl

\bibitem{mouton2011unbiased}
P.~R. Mouton, \emph{Unbiased stereology: a concise guide}.\hskip 1em plus 0.5em
  minus 0.4em\relax JHU Press, 2011.

\bibitem{mouton2017unbiased}
P.~R. Mouton, H.~A. Phoulady, D.~Goldgof, L.~O. Hall, M.~Gordon, and D.~Morgan,
  ``Unbiased estimation of cell number using the automatic optical
  fractionator,'' \emph{Journal of chemical neuroanatomy}, vol.~80, pp. A1--A8,
  2017.

\bibitem{burke2009knowing}
M.~Burke, S.~Zangenehpour, P.~R. Mouton, and M.~Ptito, ``Knowing what counts:
  unbiased stereology in the non-human primate brain,'' \emph{Journal of
  visualized experiments: JoVE}, p.~27, 2009.

\bibitem{west1991unbiased}
M.~West, L.~Slomianka, and H.~J.~G. Gundersen, ``Unbiased stereological
  estimation of the total number of neurons in the subdivisions of the rat
  hippocampus using the optical fractionator,'' \emph{The Anatomical Record},
  vol. 231, no.~4, pp. 482--497, 1991.

\bibitem{valdecasas2001extended}
A.~G. Valdecasas, D.~Marshall, J.~M. Becerra, and J.~Terrero, ``On the extended
  depth of focus algorithms for bright field microscopy,'' \emph{Micron},
  vol.~32, no.~6, pp. 559--569, 2001.

\bibitem{bradley2004one}
A.~P. Bradley and P.~C. Bamford, ``A one-pass extended depth of field algorithm
  based on the over-complete discrete wavelet transform,'' in \emph{Image and
  Vision Computing'04 New Zealand (IVCNZ'04)}.\hskip 1em plus 0.5em minus
  0.4em\relax not found, 2004, pp. 279--284.

\bibitem{hinton2006reducing}
G.~E. Hinton and R.~R. Salakhutdinov, ``Reducing the dimensionality of data
  with neural networks,'' \emph{science}, vol. 313, no. 5786, pp. 504--507,
  2006.

\bibitem{nair2010rectified}
V.~Nair and G.~E. Hinton, ``Rectified linear units improve restricted boltzmann
  machines,'' in \emph{Proceedings of the 27th international conference on
  machine learning (ICML-10)}, 2010, pp. 807--814.

\bibitem{srivastava2014dropout}
N.~Srivastava, G.~Hinton, A.~Krizhevsky, I.~Sutskever, and R.~Salakhutdinov,
  ``Dropout: A simple way to prevent neural networks from overfitting,''
  \emph{The Journal of Machine Learning Research}, vol.~15, no.~1, pp.
  1929--1958, 2014.

\bibitem{ioffe2015batch}
S.~Ioffe and C.~Szegedy, ``Batch normalization: Accelerating deep network
  training by reducing internal covariate shift,'' \emph{arXiv preprint
  arXiv:1502.03167}, 2015.

\bibitem{lecun2015deep}
Y.~LeCun, Y.~Bengio, and G.~Hinton, ``Deep learning,'' \emph{nature}, vol. 521,
  no. 7553, p. 436, 2015.

\bibitem{litjens2017survey}
G.~Litjens, T.~Kooi, B.~E. Bejnordi, A.~A.~A. Setio, F.~Ciompi, M.~Ghafoorian,
  J.~A. van~der Laak, B.~van Ginneken, and C.~I. S{\'a}nchez, ``A survey on
  deep learning in medical image analysis,'' \emph{Medical image analysis},
  vol.~42, pp. 60--88, 2017.

\bibitem{krizhevsky2012imagenet}
A.~Krizhevsky, I.~Sutskever, and G.~E. Hinton, ``Imagenet classification with
  deep convolutional neural networks,'' in \emph{Advances in neural information
  processing systems}, 2012, pp. 1097--1105.

\bibitem{ronneberger2015u}
O.~Ronneberger, P.~Fischer, and T.~Brox, ``U-net: Convolutional networks for
  biomedical image segmentation,'' in \emph{International Conference on Medical
  image computing and computer-assisted intervention}.\hskip 1em plus 0.5em
  minus 0.4em\relax Springer, 2015, pp. 234--241.

\bibitem{saper1996any}
C.~B. Saper, ``Any way you cut it: a new journal policy for the use of unbiased
  counting methods,'' \emph{Journal of Comparative Neurology}, vol. 364, no.~1,
  pp. 5--5, 1996.

\bibitem{santacruz2005tau}
K.~Santacruz, J.~Lewis, T.~Spires, J.~Paulson, L.~Kotilinek, M.~Ingelsson,
  A.~Guimaraes, M.~Deture, M.~Ramsden, E.~McGowan \emph{et~al.}, ``Tau
  suppression in a neurodegenerative mouse model improves memory function,''
  \emph{Science}, vol. 309, no. 5733, pp. 476--481, 2005.

\bibitem{spires2006region}
T.~L. Spires, J.~D. Orne, K.~SantaCruz, R.~Pitstick, G.~A. Carlson, K.~H. Ashe,
  and B.~T. Hyman, ``Region-specific dissociation of neuronal loss and
  neurofibrillary pathology in a mouse model of tauopathy,'' \emph{The American
  journal of pathology}, vol. 168, no.~5, pp. 1598--1607, 2006.

\bibitem{savitzky1964smoothing}
A.~Savitzky and M.~J. Golay, ``Smoothing and differentiation of data by
  simplified least squares procedures.'' \emph{Analytical chemistry}, vol.~36,
  no.~8, pp. 1627--1639, 1964.

\bibitem{chollet2015keras}
F.~Chollet \emph{et~al.}, ``Keras,'' \url{https://github.com/fchollet/keras},
  2015.

\bibitem{tensorflow2015-whitepaper}
\BIBentryALTinterwordspacing
M.~Abadi, A.~Agarwal, P.~Barham, E.~Brevdo, Z.~Chen, C.~Citro, G.~S. Corrado,
  A.~Davis, J.~Dean, M.~Devin, S.~Ghemawat, I.~Goodfellow, A.~Harp, G.~Irving,
  M.~Isard, Y.~Jia, R.~Jozefowicz, L.~Kaiser, M.~Kudlur, J.~Levenberg,
  D.~Man\'{e}, R.~Monga, S.~Moore, D.~Murray, C.~Olah, M.~Schuster, J.~Shlens,
  B.~Steiner, I.~Sutskever, K.~Talwar, P.~Tucker, V.~Vanhoucke, V.~Vasudevan,
  F.~Vi\'{e}gas, O.~Vinyals, P.~Warden, M.~Wattenberg, M.~Wicke, Y.~Yu, and
  X.~Zheng, ``{TensorFlow}: Large-scale machine learning on heterogeneous
  systems,'' 2015, software available from tensorflow.org. [Online]. Available:
  \url{https://www.tensorflow.org/}
\BIBentrySTDinterwordspacing

\bibitem{kinga2015method}
D.~Kinga and J.~B. Adam, ``A method for stochastic optimization,'' in
  \emph{International Conference on Learning Representations (ICLR)}, 2015.

\bibitem{simard2003best}
P.~Y. Simard, D.~Steinkraus, J.~C. Platt \emph{et~al.}, ``Best practices for
  convolutional neural networks applied to visual document analysis.'' in
  \emph{ICDAR}, vol.~3, 2003, pp. 958--962.

\end{thebibliography}
\end{document}